\documentclass{article}

\usepackage[preprint]{neurips_2026}
\usepackage{graphicx}
\usepackage{amsmath}

\usepackage[utf8]{inputenc} 
\usepackage[T1]{fontenc}    
\usepackage{hyperref}       
\usepackage{url}            
\usepackage{booktabs}       
\usepackage{amsfonts}       
\usepackage{nicefrac}       
\usepackage{microtype}      
\usepackage{xcolor}         
\usepackage{multicol}
\usepackage{multirow}

\title{Leviathan: Decoupling Input and Output Representations in Language Models}

\newcommand{\se}[1]{\textcolor{gray}{\scriptsize \,#1}}

\author{%
  Reza T. Batley\\
  Kevin T. Crofton Department of Aerospace and Ocean Engineering\\
  Virginia Polytechnic Institute and State University\\
  Blacksburg, VA 24060 \\
  \texttt{rezabatley@vt.edu} \\
  \And
  Sourav Saha \\
Kevin T. Crofton Department of Aerospace and Ocean Engineering\\
  Virginia Polytechnic Institute and State University\\
  Blacksburg, VA 24060 \\
  \texttt{souravsaha@vt.edu} \\
}

\begin{document}

\maketitle

\begin{abstract}
    Modern language models use a single matrix for input embedding and output projection. This couples two distinct objectives: token representation and discrimination over a vocabulary. This work introduces \emph{Leviathan}, a Transformer architecture that replaces the input embedding matrix with \emph{learned embedding vectorization (LEV)}, a compact continuous mapping from token indices to embeddings. Leviathan's output head remains untied for a parameter increase of as low as 0.2\%. Under controlled comparisons with identical Transformer backbones, Leviathan consistently improves language modeling performance over standard tied-embedding baselines across a 200M-1.2B parameter regime on The Pile with gains that grow during training. At 1.2B scale, Leviathan reduces validation perplexity by 9\%, requires $2.1\times$ fewer training tokens to reach the tied baseline's final loss, and improves on all six downstream benchmarks evaluated, including a 30\% reduction in LAMBADA perplexity. Frequency-stratified analysis reveals gains to be concentrated in rare tokens, where continuous parameterization reduces perplexity by 81\%, falling to near zero for the most frequent.
\end{abstract}

\section{Introduction}

Modern language models represent tokens using a shared embedding matrix serving two distinct roles: mapping input tokens into continuous representations and projecting hidden states back into vocabulary space for prediction. In most architectures, these roles are coupled through weight tying \citep{press2017tying, inan2017tying}. This forces a single matrix to simultaneously support both input representation and output classification.

These objectives impose fundamentally different requirements. This work hypothesizes that input representations benefit from smoothness, with semantically related tokens occupying nearby regions in representation space. The output head, however, is a classifier that separates all tokens across the vocabulary. Coupling these roles constrains the model to compromise in a way that may not be optimal for either objective. The most natural remedy is to untie the embeddings entirely. This, however, proves ineffective: a fully untied model underperforms the tied-weight baseline throughout training over 8.4B tokens, despite a roughly 50\% increase in parameter count. The additional parameters do not appear to be utilized effectively.

This work posits that this coupling is an architectural limitation requiring a different solution, and introduces \emph{Leviathan}. Leviathan is a Transformer architecture built around a \emph{learned embedding vectorization} (LEV) layer. This replaces the input embedding matrix with a learned function that imposes smoothness over token space as an inductive bias. Leviathan retains a fully expressive untied output head whilst decoupling input representation from output prediction. Because the LEV layer replaces the dense input embedding matrix at a fraction of its size, the untied output head adds no net parameter cost; at 1.2B scale, this overhead is 0.2\%.

Across language models from 200M to 1.2B parameters, Leviathan consistently improves validation loss over standard tied-embedding baselines. At 1.2B scale, Leviathan achieves an improvement of 0.087 nats, corresponding to 9.2\% in perplexity. This translates to improvements on downstream tasks across six tested benchmarks, including a 30\% reduction in perplexity on LAMBADA \cite{paperno2016lambadadatasetwordprediction}. These results suggest that the inductive bias of smooth input representations is more effective than decoupling alone, and that the resulting freed capacity benefits the model without the optimization cost of fully untied embeddings. Mechanistic analysis reveals the benefit is concentrated where a limited training signal makes discrete embeddings difficult to learn reliably -- across the long tail of the token distribution. For tokens appearing fewer than one-in-ten-million times in The Pile, Leviathan reduces perplexity by 81\%; for those occurring more than one-in-ten-thousand, this falls below 3\%. Continuous parameterization is most valuable precisely where discrete lookups are weakest.

\section{Background}

For a vocabulary size $V$ and hidden dimension $D$, tied-weight language models use an embedding matrix $E\in \mathbb{R}^{V\times D}$ coupling two roles: representation for inputs and discrimination for outputs. Prior work has primarily focused on reducing the parameter cost of embeddings. Factorization methods, such as ALBERT \citep{lan2020albert}, decompose the embedding matrix into lower-rank components -- reducing parameter count but retaining a linear dependence on vocabulary size $V$ and a shared input-output geometry. Another approach is to restructure the embedding space through hashing \citep{svenstrup2017hash}. These methods reduce the parameter cost of the input embedding but retain the coupled input--output formulation and do not address the tension between input representation and output classification. Character-level and subword encoding approaches \citep{kim2015character, peters2018deepcontextualizedwordrepresentations} introduce a form of input-output decoupling through a separate encoder, but at substantially greater architectural complexity. Operating at character granularity increases effective sequence length by a factor of four or more, compounding the quadratic cost of self-attention. This makes such approaches impractical at the vocabulary scales and sequence lengths used in modern language models.

In this work, the coupling itself is targeted. Prior work suggests that the output classifier benefits from being decoupled from the input embedding. Adaptive softmax \cite{baevski2019adaptive} and mixture-of-softmaxes \cite{yang2018breaking} improve output expressivity by relaxing the tied-weight constraint without addressing the input representation. By replacing the discrete input embedding with a compact continuous generator, the cost of input representation is reduced sufficiently to allow a fully expressive untied output head. This decoupling enables the model to allocate capacity to the output classifier where it is most effective, without incurring the parameter overhead of untied embeddings or the optimization difficulties that weight tying was originally introduced to avoid \cite{press2017tying, inan2017tying}. Furthermore, discrete embedding matrices are known to exhibit anisotropic and degenerate geometries under long-tailed token distributions \cite{gao2019representationdegenerationproblemtraining}. This collapse is mitigated by the continuous parameterization of the LEV layer, which shares parameters across structurally similar tokens.

\section{Method}

\subsection{Learned embedding vectorization}
\label{sec:LEV}

The continuous embedding layer $\mathcal{G}:\{0,\dots,V-1\}\to \mathbb{R}^D$ maps token indices to embeddings. Each token index $i$ is mapped to a factorized index via a compositional indexing scheme. The vocabulary is factorized into $k$ components using a base-$b$ decomposition, producing indices $(i_1,\dots,i_k)$. These index into shared codebooks $C_1,\dots,C_k\in\mathbb{R}^{b\times d_{\mathrm{seed}}}$ to yield a seed representation $z(i)=\sum_{r=1}^kC_r[i_r]$. The base-$b$ decomposition serves as a parameter-sharing mechanism, through which rare tokens benefit from gradient signal passed through shared codebook entries.

The seed is projected independently for each of $h$ heads. For a given head $\ell$, this projection is normalized to a bounded latent coordinate $\tilde z_\ell =\sigma\left(\mathrm{LN}\left(\frac 1 2 W_{\mathrm{seed},\ell}z(i)\right)\right)\in [0,1]^{d_{\mathrm{seed}}}$ which is passed through a learned separable function. The scaling factor $\frac 1 2$ is stabilizing: sigmoid inputs are kept near the linear regime. Each dimension is modeled by a univariate quadratic B-spline basis expansion with $\kappa$ knots, and combined by a rank-$r$ tensor product.

Embeddings are obtained by summing across heads after a final linear projection to the transformer's hidden width, $E_i=\sum_{\ell=1}^hW_\ell\mathcal{M}_\ell(\tilde z_\ell)$, $W_\ell\in\mathbb{R}^{r\times D}$. Parameters are shared across tokens through the continuous mapping, enabling compact representations. An overview of this component is shown in Figure \ref{fig:model} (\emph{left}). 

The LEV layer has a parameter count of $kbd_\mathrm{seed}+h(d^2_\mathrm{seed}+2d_\mathrm{seed}+d_\mathrm{seed}\kappa r+d_\mathrm{seed}+rD)$. Furthermore, it requires $2h(d^2_\mathrm{seed}+d_\mathrm{seed}\kappa r + rD)$ FLOPs per token -- 4.2M at 1.2B scale. This represents $<0.3\%$ overhead with the Transformer and output projection requiring 1.6B FLOPs per token.

For the main results, the \texttt{o200k\_base} tokenizer is used with $V=200,376$. Setting $k=3$ and $b=59$ yields a representable vocabulary of $59^3=205,379$, which covers the tokenizer with minimal unused capacity. Unless stated otherwise, models use $h=8, r=64, \kappa=16$ and $d_\mathrm{seed}=128$. This corresponds to 1.21M+$512D$ parameters. At $D=2048$, this is 2.25M parameters compared to $V\cdot D=410.4M$ for a dense input table.

\subsection{Model architecture}

\begin{figure}
    \centering
    \includegraphics[width=0.55\linewidth]{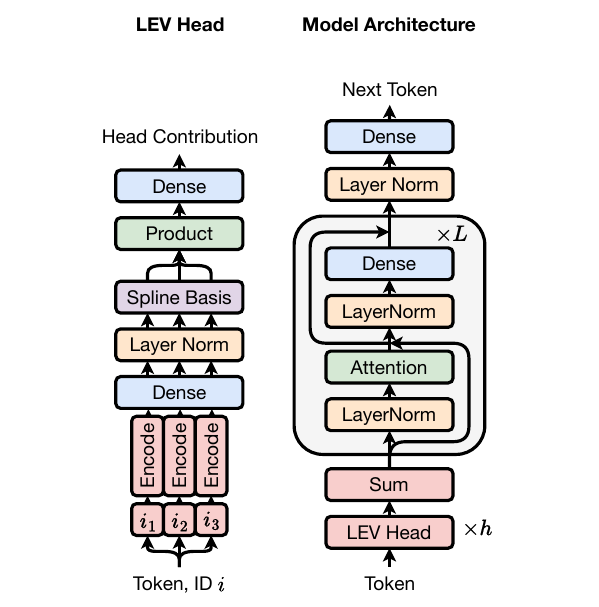}
    \caption{\emph{Left}, LEV head: maps token index $i$ to an embedding via compositional base-$b$ encoding, a learned seed projection, separable B-spline basis expansion and rank-$r$ tensor product. \emph{Right}, The full Leviathan architecture. $h$ LEV heads are summed to produce the input embedding, which is fed into a standard $L$-layer decoder-only Transformer.}
    \label{fig:model}
\end{figure}

Embeddings are fed into a decoder-only Transformer with hidden dimension $D$, $L$ layers and $H$ self-attention heads. To isolate the effects of the LEV layer, a standard backbone (\ref{sec:setup}) is used across all models. The final hidden states are projected to the vocabulary using a dense output head. In contrast to tied-weight models, the LEV layer enables this output head to remain fully untied for a less than 0.2\% parameter overhead. A schematic of the full architecture is shown in Figure \ref{fig:model} (\emph{right}).

\section{Experimental setup}
\label{sec:setup}

\paragraph{Models}

All models use a decoder-only Transformer backbone with pre-layer normalization, rotary positional embeddings \cite{su2023rope} and SwiGLU feedforward layers \cite{shazeer2020glu}. Three model scales are evaluated: 200M, 400M and 1.2B parameters. Architecture dimensions are summarized in Table \ref{tab:models}. At each scale, the tied-weight baseline (\textsc{Tied}) and Leviathan share an identical backbone, differing only in input--output parameterization. Parameter counts are reported inclusive of all embedding matrices. All models are trained with identical optimization hyperparameters, seed, data order and compute budget at each scale.

\begin{table}[!h]
    \centering
    \begin{tabular}{lcccccccc}
    \toprule
         \multirow{2}{*}{Scale} & \multirow{2}{*}{$D$} & \multirow{2}{*}{$L$} & \multirow{2}{*}{$H$} & \multicolumn{2}{c}{Parameters} & \multicolumn{2}{c}{LEV Overhead} & \multirow{2}{*}{Tokens} \\
         \cmidrule(lr){5-6}\cmidrule(lr){7-8}
         & & & & Tied & Leviathan & Params, \% & FLOPs, \% \\
         \midrule
         200M & 640 & 16 & 5 & 206,946,560 & 208,679,992 & $0.84\%$ & $0.82\%$ & 4.2B \\
         400M & 1024 & 16 & 8 & 406,611,968 & 408,542,008 & $0.48\%$ & $0.51\%$ & 8.4B \\
         1.2B & 2048 & 16 & 16 & 1,215,909,888 & 1,218,364,216 & $0.20\%$ & $0.26\%$ & 24.1B\\
         \bottomrule
    \end{tabular}
    \vspace{1pt}
    \caption{Model configurations, parameters and token budgets.}
    \label{tab:models}
\end{table}

\paragraph{Experiments}

Primary comparisons evaluate \textsc{Tied} and Leviathan at 200M, 400M and 1.2B with the \texttt{o200k\_base} tokenizer \cite{tiktoken2023}, matching modern tokenizer regimes. At 400M, two additional baselines are included: (i) a fully untied model ($\approx$611M parameters) serving as an upper bound on input--output decoupling at higher parameter and optimization cost. (ii) an input-parameter-matched ALBERT-style factorization \cite{lan2020albert}, paired with the same untied output head, serves as a linear compression baseline. Where Leviathan generates input embeddings via a learned nonlinear function, the factorized embedding decomposes the input embedding matrix into a low-rank product -- isolating the contribution of nonlinearity in the input representation.

To study the effect of vocabulary size, 400M experiments are repeated with \texttt{GPT-2} ($V=$50,257) and \texttt{cl100k\_base} ($V=$100,277) tokenizers, reducing model parameters to $\approx$250M and $\approx$300M respectively. All primary experiments use \texttt{o200k\_base}; vocabulary ablations are conducted at 400M only, with the consistency of results suggesting that $V$ is not a significant moderator of the effect.

Mechanistic ablations at 400M replace the LEV layer with (i) a parameter-matched two-layer SiLU MLP and (ii) a parameter-matched factored MLP that preserves the input decomposition but removes the separable B-spline structure. The flat MLP ablation measures how much of Leviathan's improvement comes from continuous parameterization alone; the factored MLP ablation isolates the additional contribution of the separable B-spline structure.

\paragraph{Training}

All models are trained on The Pile \cite{gao2020pile} (uncopyrighted), using a context length of 512 tokens and an effective batch size of 512 sequences (262,144 tokens/step) via gradient accumulation over 2 steps. These are held fixed to prevent confounding across scales. The optimizer is AdamW \cite{loshchilov2018decoupled} with $\beta_1=0.9,\beta_2=0.999,\varepsilon=10^{-8}$, and no weight decay to avoid confounding regularization effects across embedding parameterizations. The learning rate follows a warmup-cosine schedule with linear warmup from $10^{-5}$ to the peak value, decaying to 10\% of the peak at the final step. Peak learning rates are $4\times10^{-4}$ at 200M, $3\times10^{-4}$ at 400M and $2\times10^{-4}$ at 1.2B. Warmup spans the first 320 steps at 200M, 640 at 400M and 1,000 steps at 1.2B. Gradients are clipped to unit norm. All models are trained in bfloat16 with float32 master weights on a single NVIDIA H200 GPU.

\paragraph{Evaluation protocol}

Language modeling performance is evaluated on a held-out validation shard of The Pile (200M and 400M use shard 0 of 1,024 from \texttt{monology/pile-uncopyrighted}, 1.2B use the \texttt{EleutherAI/pile} validation split). Zero-shot generalization to Wikitext-103 \cite{merity2018scalable} is evaluated on the test split. The corpus is frozen, then segmented into 555 non-overlapping windows of 512 tokens totaling 283,605 tokens predicted. Boundary effects are consistent across all models. Downstream task performance is measured zero-shot using \texttt{lm-evaluation-harness} \cite{gao2024harness} on HellaSwag \cite{zellers2019hellaswagmachinereallyfinish}, ARC Easy and Challenge \cite{clark2018thinksolvedquestionanswering}, PIQA \cite{bisk2019piqareasoningphysicalcommonsense}, WinoGrande \cite{sakaguchi2019winograndeadversarialwinogradschema} and LAMBADA \cite{paperno2016lambadadatasetwordprediction}. HellaSwag, ARC and PIQA report normalized accuracy; WinoGrande and LAMBADA report standard accuracy.

\section{Results}

\subsection{Language modeling performance}

\begin{figure}[!t]
    \centering
    \includegraphics[width=\linewidth]{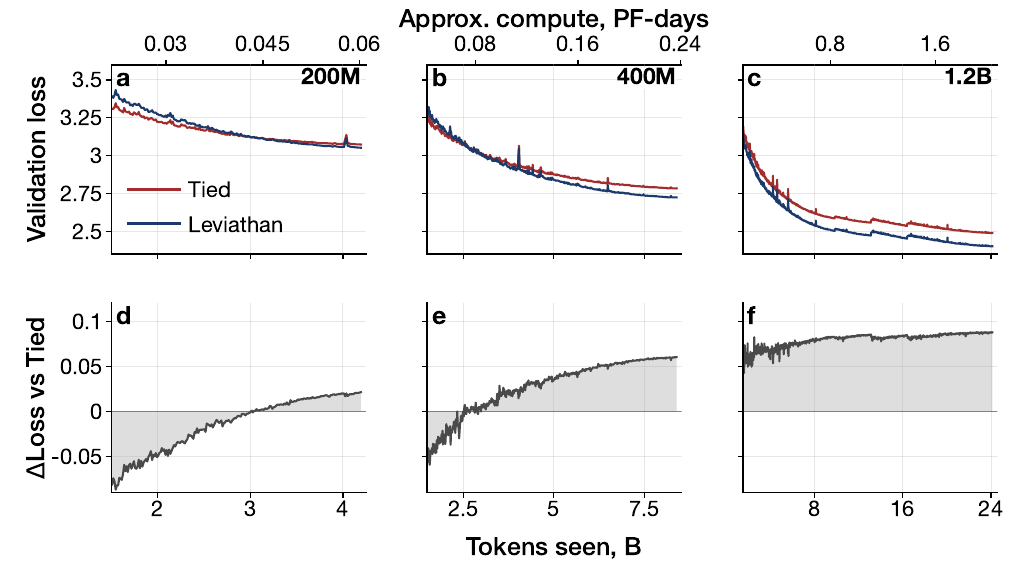}
    \caption{\textbf{a--c}, Validation loss curves with respect to billions of tokens seen and PF-days of compute ($\pm1\%$) on The Pile for Leviathan (navy) and Tied (crimson) at 200M, 400M and 1.2B.\textbf{d--f}, Corresponding loss improvement (Tied $-$ Leviathan, nats).}
    \label{fig:main}
\end{figure}

Leviathan is evaluated against a tied-embedding baseline under controlled conditions: identical Transformer backbones, training data, tokenizer and hyperparameters, only varying the input representation. Results are reported from a fixed held-out validation set from The Pile across model sizes. Leviathan consistently improves validation loss over the tied baseline across all scales with an advantage that grows throughout training (Figure \ref{fig:main}). 

At 200M, Leviathan achieves a final validation loss of 3.052 nats compared to 3.073 for \textsc{Tied}, a reduction of 0.7\% in loss and 2.1\% in perplexity over 4.2B tokens. At 400M, the gap widens to 2.2\% in loss (5.6\% in perplexity) over 8.4B tokens, with Leviathan reaching 2.724 nats against 2.784 nats. At 1.2B, the gap widens to $3.5\%$ in loss ($9.2\%$ in perplexity), Leviathan reaching 2.402 nats compared to 2.489 nats over 24.1B tokens. 

At all scales, the advantage grows monotonically throughout training, suggesting that the benefit compounds with additional data. At 200M, Leviathan initially underperforms the baseline before crossing over at approximately 3B (Figure \ref{fig:main}\textbf{d}). This early optimization disadvantage diminishes with scale, and is absent at 1.2B (Figure \ref{fig:main}\textbf{f}).  

\subsection{Comparison to untied embeddings}

\begin{figure}[!t]
    \centering
    \includegraphics[width=\linewidth]{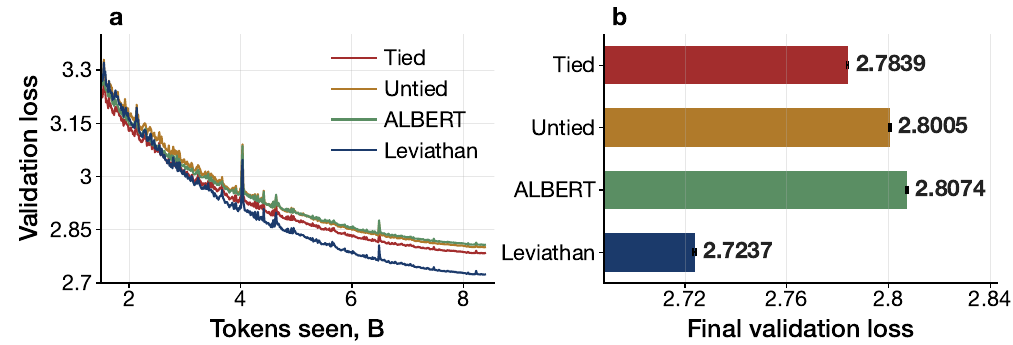}
    \caption{\textbf{a}, Validation loss at 400M for \textsc{Tied}, \textsc{Untied}, \textsc{Albert} and Leviathan. \textbf{b}, Final validation loss for each model. Error bars denote $\pm1$ standard error taken over the final 10\% of steps.}
    \label{fig:untied}
\end{figure}

The natural alternative to tied embeddings is to learn separate input and output matrices, fully untying the model at the cost of an additional $V\times D$ parameters. At 400M scale, this yields a model of 611M parameters. Despite this 50\% parameter increase, a fully untied model \textsc{Untied} underperforms the tied baseline throughout training; it reaches a final validation loss of 2.801 nats compared to 2.784 for \textsc{Tied} (Figure \ref{fig:untied}). The performance gap closes at a decelerating rate -- from $-0.178$ nats at 0.26B tokens to $-0.017$ nats at 8.4B tokens. This suggests that learning two independent dense embedding matrices over a 200k vocabulary presents an optimization difficulty that is not resolved within a Chinchilla-optimal token budget \cite{hoffmann2022training}. These additional parameters are not efficiently utilized within the studied compute regime.

The contribution of low-rank linear compression is isolated by evaluating \textsc{Albert}. This model has factorized input embeddings -- parameter-matched to the LEV layer -- paired with a full, untied unembedding head. \textsc{Albert} reaches a final 2.807 nats, worse than both \textsc{Tied} and \textsc{Untied}. Linear compression of the input embedding without nonlinear, continuous structure thus fails to recover even the tied baseline, suggesting that factorization alone disrupts the input geometry without providing compensating representational benefit.

In contrast, Leviathan crosses the tied baseline at approximately 2.5B tokens and reaches a final validation loss of 2.724, outperforming \textsc{Tied}, \textsc{Untied} and \textsc{Albert}. Leviathan achieves this with a parameter overhead of under 0.5\% relative to \textsc{Tied}, two-thirds of the parameters of \textsc{Untied} and at the same parameters as \textsc{Albert}. This demonstrates that the benefit of input--output decoupling depends on how the input representation is parameterized. Nonlinear continuous generation outperforms both naive untying and linear factorization within a Chinchilla-optimal compute budget \cite{hoffmann2022training}.

\begin{table*}[!h]
\centering
\caption{Zero-shot downstream accuracy (\%) across all models and scales.
HellaSwag, ARC-E, ARC-C, and PIQA report normalised accuracy;
WinoGrande and LAMBADA report standard accuracy. Standard errors in parentheses. \textbf{Bold} indicates best result within each scale group.}
\label{tab:downstream}
\resizebox{\textwidth}{!}{
\begin{tabular}{llcccccc}
\toprule
Scale & Model & HellaSwag & ARC-E & ARC-C & PIQA & WinoGrande & LAMBADA \\
\midrule
\multirow{2}{*}{200M}
 & \textsc{Tied}
   & 26.8\se{(0.4)} & \textbf{32.8\se{(1.0)}} & \textbf{22.0\se{(1.2)}}
   & 56.2\se{(1.2)} & \textbf{52.9\se{(1.4)}} & \textbf{15.7\se{(0.5)}} \\
 & \textsc{Leviathan}
   & 26.8\se{(0.4)} & 32.3\se{(1.0)} & 21.9\se{(1.2)}
   & \textbf{57.5\se{(1.2)}} & 49.8\se{(1.4)} & 14.5\se{(0.5)} \\
\midrule
\multirow{2}{*}{400M}
 & \textsc{Tied}
   & 27.9\se{(0.4)} & 35.2\se{(1.0)} & 22.1\se{(1.2)}
   & 59.0\se{(1.1)} & \textbf{51.1\se{(1.4)}} & \textbf{23.9\se{(0.6)}} \\
 & \textsc{Leviathan}
   & \textbf{28.4\se{(0.5)}} & \textbf{39.0\se{(1.0)}} & \textbf{22.9\se{(1.2)}}
   & \textbf{61.0\se{(1.1)}} & 50.0\se{(1.4)} & 23.5\se{(0.6)} \\
\midrule
\multirow{2}{*}{1.2B}
 & \textsc{Tied}
   & 31.7\se{(0.5)} & 38.7\se{(1.0)} & 23.1\se{(1.2)}
   & 61.3\se{(1.1)} & 48.5\se{(1.4)} & 32.6\se{(0.7)} \\
 & \textsc{Leviathan}
   & \textbf{33.2\se{(0.5)}} & \textbf{42.6\se{(1.0)}} & \textbf{23.5\se{(1.2)}}
   & \textbf{63.5\se{(1.1)}} & \textbf{51.5\se{(1.4)}} & \textbf{33.4\se{(0.7)}} \\
\bottomrule
\end{tabular}}
\end{table*}

\begin{figure}[!t]
    \centering
    \includegraphics[width=\linewidth]{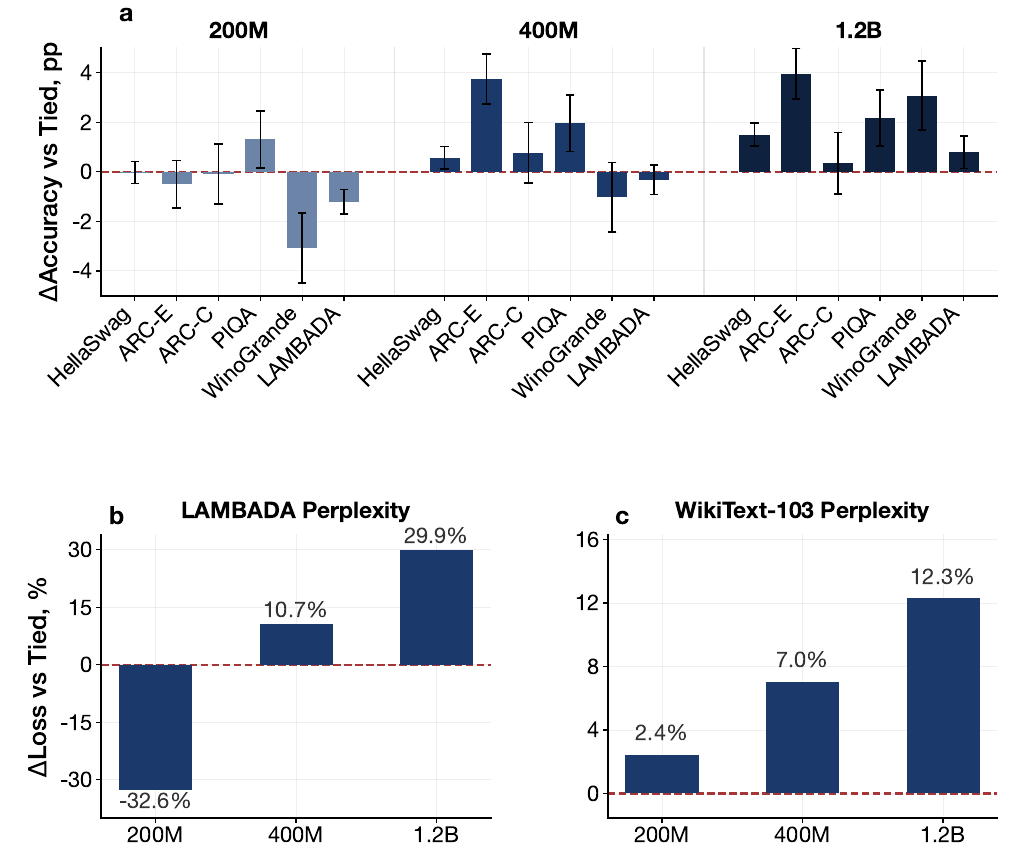}
    \caption{\textbf{a}, Zero-shot accuracy gain of Leviathan over \textsc{Tied} (pp) across six benchmarks at 200M, 400M and 1.2B scale; error bars $\pm$1 standard error. \textbf{b}, LAMBADA perplexity improvement (\%) vs \textsc{Tied}. \textbf{c}, WikiText-103 zero-shot transfer perplexity improvement vs \textsc{Tied}.}
    \label{fig:downstream}
\end{figure}

\subsection{Generalization and downstream performance}

Leviathan is evaluated zero-shot on six downstream benchmarks, shown in full in Table \ref{tab:downstream}. At 200M, downstream performance is mixed: Leviathan generally matches \textsc{Tied}, outperforming it on PIQA but underperforming on WinoGrande and LAMBADA. At 400M, gains emerge with Leviathan matching \textsc{Tied} within one standard error, except on ARC-Easy and PIQA with +3.7pp and +2.0pp improvements, respectively. At 1.2B, Leviathan improves on all six tasks, including WinoGrande (+3.1pp) with a mean accuracy gain of 2.3pp across the benchmarks. These results are visualized in Figure \ref{fig:downstream}\textbf{a}.

Figure \ref{fig:downstream}\textbf{b} shows a pronounced scaling trend: despite lagging 33\% at 200M, Leviathan reduces LAMBADA perplexity by 11\% at 400M and 30\% at 1.2B relative to \textsc{Tied}. Zero-shot transfer to WikiText-103 \cite{merity2018scalable} -- a held-out corpus from a different distribution, evaluated without adaptation -- shows improvements of 2.4\%, 7.0\% and 12.3\% in perplexity at 200M, 400M and 1.2B, respectively (Figure \ref{fig:downstream}\textbf{c}). The increased advantage of Leviathan with scale suggests that its continuous input representations learn token geometries that more effectively transfer across data distributions than discrete lookup tables.

\subsection{Effect of vocabulary size}

All primary experiments use the \texttt{o200k\_base} tokenizer with $V=200{,}376$, a large vocabulary reflecting modern tokenizer regimes. To verify that results are not specific to this choice, 400M-scale experiments are repeated with \texttt{GPT-2}  ($V=50{,}257)$ and \texttt{cl100k\_base} ($V=100{,}277$) tokenizers. This reduces total model parameters to around 250M and 300M respectively as their embedding matrices shrink. Across these settings, this matrix represents 20\%, 30\% and 50\% of total model parameters. The only modification to the LEV layer is taking $b=37$ at 50k, as $37^3=50{,}653$ and $b=47$ at 100k, as $47^3=103{,}823$.

\begin{figure}[!h]
    \centering
    \includegraphics[width=\linewidth]{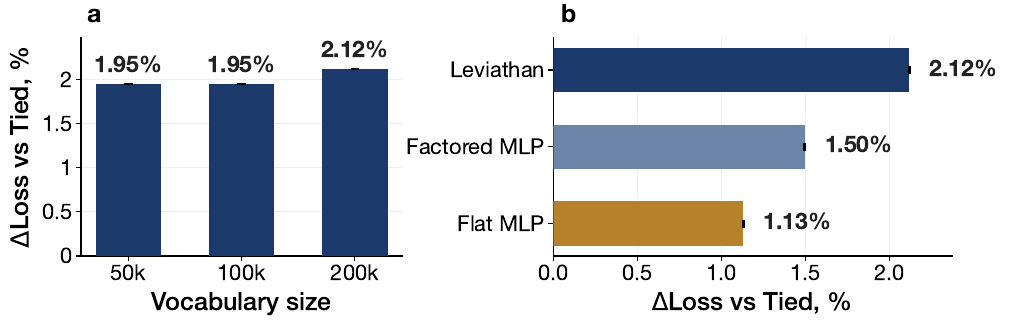}
    \caption{\textbf{a}, Validation loss improvement of Leviathan over \textsc{Tied} (\% relative) across vocabulary sizes at 400M scale. \textbf{b}, Mechanistic ablation at 400M scale, measuring improvement of architectural modifications of the LEV layer compared with \textsc{Tied}. All error bars denote $\pm1$ standard error taken over the final 10\% of steps.}
    \label{fig:ablation}
\end{figure}

As shown in Figure \ref{fig:ablation}\textbf{a}, Leviathan achieves improvements of 1.95\%, 1.95\% and 2.12\% over \textsc{Tied} at 50k, 100k and 200k vocabularies respectively. All are within a single standard error of each other. The stability of this result thus holds over a fourfold range in vocabulary across settings where the embedding represents between one-fifth and one-half of all model parameters. This suggests that the benefit of the LEV layer is not sensitive to vocabulary scale or to the relative cost of input representation.

\subsection{Mechanistic ablation study}

To isolate the contributions of individual architecture choices within the LEV layer, two intermediate ablations are evaluated at 400M (Figure \ref{fig:ablation}\textbf{b}). First, the LEV layer is replaced with a parameter-matched two-layer SiLU MLP (\textsc{Flat MLP}), which generates input embeddings continuously and with a nonlinearity but without compositional structure or spline parameterization. Second, a parameter-matched factorized MLP (\textsc{Factored MLP}) preserves the base-$b$ token decomposition of the LEV layer, but replaces the separable B-spline basis.

\textsc{Flat MLP} improves over \textsc{Tied} by 1.13\%, establishing that continuous input parameterization with an untied output head is beneficial, even without compositional or spline structure. \textsc{Factored MLP} improves further to 1.50\%, demonstrating that the compositional indexing scheme, which encourages the sharing of representations across structurally similar tokens, contributes independently of the spline structure. Leviathan achieves the highest improvement at 2.13\%, with the remaining gain attributable to the separable B-spline parameterization, imposing smoothness and tensor-decomposed structure as an explicit bias over the bounded latent coordinate space. Each component of the LEV layer thus contributes incrementally. Continuous representation over flat lookup (+1.13\%), compositional structure over flat continuous (+0.37\%) and separable spline smoothness over unstructured factorization (+0.62\%). 

\subsection{Frequency-dependent generalization}

Leviathan parameterizes input embeddings via a continuous function over a factorized token index space. Such an approach shares parameters across tokens with overlapping base-$b$ components -- an inductive bias toward smooth interpolation. Neighboring tokens in index space share gradient signal during training. The benefit of this sharing should therefore be largest for \emph{rare} tokens: a lookup table receives insufficient gradient signal to converge to a reliable representation.

\begin{figure}[!h]
    \centering
    \includegraphics[width=\linewidth]{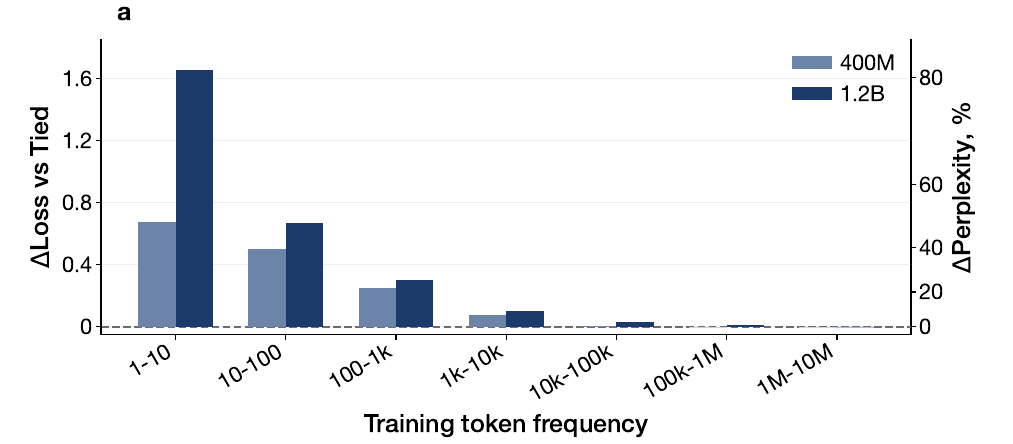}
    \caption{Per-frequency-bucket nat improvement of Leviathan over \textsc{Tied} at 400M and 1.2B scale, left axis: Loss improvement, nats; right axis: Perplexity improvement, \%. Buckets are right-exclusive.}
    \label{fig:frequency}
\end{figure}

To test this hypothesis, the vocabulary is partitioned into frequency buckets based on co-occurrence counts in 100M tokens of the training corpus, and per-bucket mean negative log-likelihoods (NLL) are computed for both \textsc{Tied} and Leviathan at 400M and 1.2B scale. Buckets range from 2,325 evaluation tokens (1--9 occurrences) to 514,653 (10k--100k); although rare-token buckets are smaller, they are sufficiently populated for stable NLL estimates. Results are shown in Figure \ref{fig:frequency}.

The frequency gradient is monotonic; at 400M, tokens seen 1--9 times in training improve by 0.68 nats ($49\%$ perplexity reduction) relative to \textsc{Tied}. The improvement decays across frequency decades: 0.50 nats for tokens seen 10--99 times, 0.25 nats for 100-999 and 0.08 nats for 1k-10k occurrences. For tokens seen more than 10k times the difference is less than 0.01 nats (0.007 for 10k--100k, -0.004 for 100k--1M, -0.007 for 1M--10M).

At 1.2B scale, the gradient steepens substantially. Rare tokens, 1--9 occurrences, improve by 1.65 nats -- an 81\% perplexity reduction, and a $2.4\times$ amplification versus 400M. The improvement decays: from 10--99, 0.67 nats, from 100--999, 0.30 nats, from 1k--10k, 0.10 nats, from 10k--100k, 0.028 nats, at 100k--1M, 0.012 nats. At 1M--10M the difference is negative and it is negligible: -0.001 nats.

This pattern provides evidence for a mechanism underlying the aggregate scaling result. The overall difference in nats grows from 0.060 at 400M to 0.087 nats at 1.2B. At 400M, gains are concentrated in the rare-token tail -- the crossover occurs between the 10k--100k and 100k--1M buckets. At 1.2B, Leviathan improves across essentially the entire vocabulary, shifting the crossover point to beyond 1M occurrences -- covering only the eight most common tokens. Only on three does Leviathan score worse. At this frequency regime, both models have received sufficient gradient signal to learn near-optimal representations; the residual differences are effectively noise around zero. This is consistent with Leviathan's smoothness inductive bias. For tokens whose optimal representations are idiosyncratic with respect to their neighbors in the factorized index space, a small bias is introduced that the tied-weight model does not share. Leviathan learns better representations specifically where discrete lookup embeddings are inefficient -- and at sufficient scale, that encompasses nearly the entire vocabulary.

\section{Discussion}

\subsection{Training overhead and inference cost}

The LEV layer introduces per-step training overhead of 26\% at 200M, 17\% at 400M and 9\% at 1.2B, diminishing with scale as the Transformer body dominates compute. At 200M, this overhead exceeds the sample efficiency gain, yielding a net wall-clock penalty of 18\%. At 400M, sample efficiency wins out: Leviathan reaches the tied baseline's final validation loss in 13\% less wall-clock time. At 1.2B, Leviathan is $2.1\times$ more sample efficient and reaches the tied baseline's final loss in half the wall-clock time. At inference, embeddings are computed once and cached in 76.8ms on a single H200 for the full 200k vocabulary. Subsequent forward passes are then identical in cost to the tied baseline. 

\subsection{Limitations}

All experiments use a single training corpus (The Pile), a single tokenizer family (BPE) and a maximum scale of 1.2B. Each configuration is trained with a single seed. Whilst this precludes formal confidence intervals at a given scale, the consistency of results across three scales, three vocabulary sizes and four architectural variants provides evidence that this effect is robust. The frequency-dependent improvement pattern is consistent with the hypothesis but is correlational. Whether gains persist at frontier scale -- where embedding parameters represent a smaller fraction of total parameters and token budgets are larger -- remains future work, although the monotonically growing gap across studied scales is encouraging. Extending to 7B+ scale and multilingual corpora would test whether the frequency-dependent mechanism generalizes.

\section{Conclusion}

This work demonstrates that the standard practice of weight-tying input and output representations imposes an architectural limitation that is not resolved by naive untying. Leviathan replaces the input embedding matrix with a compact continuous generator -- the LEV layer. This decouples input representation from output classification and imposes smoothness over token space as an inductive bias. Across 200M to 1.2B parameters, Leviathan consistently outperforms tied-weight baselines with gains that grow with scale, with data, and with training duration. Mechanistic analysis reveals that the gains are concentrated in the sparse-data regime, where continuous parameter sharing provides the largest benefit over discrete lookup embeddings. At sufficient scale, this advantage extends across nearly the entire vocabulary. These results suggest that the representational geometry of input embeddings matters, and that continuous parameterization is a data- and compute-efficient means of improving it.

\bibliographystyle{unsrt}
\bibliography{references}

\end{document}